\documentclass[biblatex, english]{lni}
\usepackage{booktabs}
\usepackage{graphicx}
\usepackage{tabularx}
\usepackage{amsfonts}
\usepackage{amsmath}
\usepackage{dsfont}

\begin{document}

\title[Privacy-preserving LM Approaches]{A Study of Privacy-preserving Language Modeling Approaches}
\author[1]{Pritilata Saha}{psaha@mail.upb.de}{0000-0002-7776-1620}
\author[1]{Abhirup Sinha}{abhirup@mail.upb.de}{0000-0002-6927-5526}
\affil[1]{Department of Computer Science\\Paderborn University\\Germany}
\maketitle

\begin{abstract}
Recent developments in language modeling have increased their use in various applications and domains. Language models, often trained on sensitive data, can memorize and disclose this information during privacy attacks, raising concerns about protecting individuals' privacy rights. Preserving privacy in language models has become a crucial area of research, as privacy is one of the fundamental human rights. Despite its significance, understanding of how much privacy risk these language models possess and how it can be mitigated is still limited. This research addresses this by providing a comprehensive study of the privacy-preserving language modeling approaches. This study gives an in-depth overview of these approaches, highlights their strengths, and investigates their limitations. The outcomes of this study contribute to the ongoing research on privacy-preserving language modeling, providing valuable insights and outlining future research directions.
\end{abstract}


\section{Introduction}
Recent work has shown that large language models (LLMs) tend to memorize information from training data containing personally identifiable information (PII), and adversaries can extract this information later \cite{carlini2021extracting, nasr2023scalable, heikkila2022does}. However, everyone has the Right to be Forgotten under the General Data Protection Regulation (GDPR) law \cite{graves2021amnesiac}.
Though current state-of-the-art LLMs perform well in generating human-like text, recent research has revealed the vulnerability of these models to preserve privacy \cite{kim2024propile, nasr2023scalable}.  

Due to the legal obligations and ethical responsibilities associated with using language models,  privacy-preserving practices are important. Recent privacy-preserving language models employ various approaches, such as Differential Privacy \cite{anil2021large, abadi2016deep, wu2022adaptive,shi2021selective,shi2022just}, Knowledge Unlearning \cite{jang2022knowledge, yao2023large}, Data Preprocessing \cite{lison-etal-2021-anonymisation, lee2021deduplicating, kandpal2022deduplicating}, Private Representation Learning \cite{zhou-etal-2022-textfusion, zhou2023textobfuscator}, Federated Learning \cite{lin2021fednlp, mcmahan2017communication, reddi2020adaptive, li2020federated} to mitigate inherent privacy risks. All the current privacy-preserving methodologies safeguard privacy against different types of attacks, but no single approach can protect against all kinds of privacy attacks alone.  

This study reviews the core concepts underlying the most used approaches for mitigating privacy risks, along with their benefits and challenges. This study is divided into four independent research directions. 
\begin{itemize}
    \item \textbf{Differential Privacy (DP)-based approaches:} Differential Privacy-based approaches apply existing DP algorithms, sometimes with some updates, e.g., DP-SGD, to protect privacy information from leaking from training data.
    \item \textbf{Private Representation Learning-based approaches:} These approaches can preserve the privacy of the representations to avoid text reconstruction attacks.
    \item \textbf{Knowledge Unlearning-based approaches:} These approaches use algorithms, e.g., negative log-likelihood, to forget specific sequences of tokens from training data.
    \item \textbf{Data Preprocessing approaches:} These approaches focus on detecting and removing sensitive information from training data to mitigate privacy risks.
\end{itemize}
This research discusses each approach's merits and shortcomings and suggests future research directions depending on them.

The remainder of this paper is structured as follows. First, it gives an overview of what privacy-preserving language modeling is (Section \ref{sec: What is preserving privacy in Language Models?}). Then, discuss the four widely used research methods and approaches (Section \ref{sec: Research Methods and Approaches}), followed by the key findings (Section \ref{sec: Key findings}) where findings of this work are added. Finally, the limitations and future works (Section \ref{sec: Limitation and Future works}) are followed by the conclusion (Section \ref{sec: Conclusion}).

\section{Why Preserve Privacy in Language Models?}
\label{sec: What is preserving privacy in Language Models?}
LLMs have recently become an integral part of our lives, and these models can be used for different tasks, e.g., text generation and language translation. These language models have been widely used in chatbots, AI assistance systems, etc. But when it comes to privacy, most of the models struggle to preserve privacy. 
\subsection{Legal Requirements}
According to the Universal Declaration of Human Rights \cite{johnson1988universal}, privacy is one of the fundamental human rights, and individuals should not face unwarranted intrusion into their privacy.
The General Data Protection Regulation (GDPR) was adopted in 2018 \cite{gdpr}, and it gives individuals control over their personal data.
Every individual has the right to limit the use of their personal information, and the Right to be Forgotten is part of the General Data Protection Regulation (GDPR) law \cite{graves2021amnesiac,mantelero2013eu}. 
Therefore, addressing privacy concerns in LMs is not only a social responsibility, but there are also legal requirements to ensure compliance with established human rights and data protection laws.

\subsection{Privacy Risks}
Most current state-of-the-art models can not ensure the privacy of personally identifiable information. Language models are trained with highly sensitive data that contains personally identifiable information (PII), such as names, email addresses, and phone numbers. These models tend to memorize the knowledge from initial pretraining, and recent works have shown that adversaries can extract training data from language models \cite{heikkila2022does}.  

Preserving privacy in language models involves implementing different approaches to mitigate privacy risks. Current privacy-preserving methodologies utilize techniques like Differential Privacy (DP), Knowledge Unlearning, Private Representation Learning, etc. These methods aim to protect the disclosure of sensitive information in the training data under privacy attacks. Section \ref{sec: Research Methods and Approaches} discusses some of such available methods in privacy-preserving NLP.

\section{Research Methods and Approaches}
\label{sec: Research Methods and Approaches}
\subsection{Differential Privacy-based approaches}
Differential privacy \cite{dwork2008differential, dwork2014algorithmic} is one of the most used approaches to preserving privacy in data. 
Differential Privacy-based approaches \cite{anil2021large, abadi2016deep, wu2022adaptive,shi2021selective,shi2022just} in language modeling aim to protect sensitive information by employing an $(\epsilon; \delta)-DP$ algorithm. An $(\epsilon; \delta)-DP$ algorithm's objective is to limit its output's use to probabilistically determine the presence of a single record in the dataset by a factor of $e^\epsilon$. DP-based approaches for preserving privacy in language models use DP-SGD optimization proposed by \citet{abadi2016deep}. The fundamental concept behind DP-SGD involves clipping each example's gradients and adding Gaussian noise $z \sim \mathcal{N}(0, C^2\sigma^2I)$ during training. The new gradient is calculated as,
\begin{align}
    \tilde{g_{L_t}} = \frac{1}{L} \left(\sum_{x_i} g(x_i) + z_t\right)
    \label{eq:regular_DP}
\end{align}
The adaptive noise technique, suggested by \citet{wu2022adaptive}, dynamically modifies the noise magnitude based on the privacy probability of an item within the DP-SGD process. Ultimately, a gradient optimization algorithm incorporating adaptive noise is presented in equation \ref{eq: adaptive DP}.
\begin{align}
    \gamma_B = \frac{\sum_{i=1}^{L}\rho(s_i)}{L}
    \label{eq: privacy weights}
\end{align}
\begin{align}
    z_{\text{Badp}} = \gamma_B \cdot \mathcal{N}(0, C^2\sigma^2I^2)
    \label{eq: adaptive DP}
\end{align}
Here, $\gamma_B$ denotes the privacy weights, which is the privacy probability averaged over batch $B = \{s_1, s_2, \ldots, s_L\}$ of size $L$ as shown in equation \ref{eq: privacy weights}. $\mathcal{N}(0, C^2\sigma^2I_2)$ is the Gaussian noise of $B$, where $\sigma$ is a noise multiplier, and $C$ is the clipping norm.

While the DP approaches mentioned earlier focus on the overall data, the Selective Differential Privacy approach proposed by \citet{shi2021selective,shi2022just} focuses on the privacy-sensitive portion of the data only to provide a privacy guarantee. This approach uses a policy function $F$ to distinguish between private and non-private attributes inside one data point and, in this way, protect the privacy of the sensitive parts while maintaining model utility.
In this approach, the policy function is first used to get the privacy bit matrix $P=F(D)$. Here, for a record $r \in \tau$, the policy function $F: \tau \mapsto \{0, 1\}^{n_r}$ finds sensitive attributes by assigning $F(r)_i = 0$ and non-sensitive attributes assigning $F(r)_i = 1$. Here, $n_r$ is the number of attributes in $r$. Policy function can be defined according to the application.
After that, the Selective-DPSGD algorithm is used to train the model. Within this algorithm, the regular stochastic gradient descent (SGD) algorithm is used for non-private updates, and DP-SGD is used for private updates. The private and non-private updates are determined by the privacy bit matrix $P$.

\paragraph{Benefits:} 
DP can provide strong privacy over the entire dataset, so it can be used to preserve the privacy of PII. Though some research suggested the performance degradation issue, the most recent approaches, e.g., Adaptive DP and Selective DP, are designed to maintain model utility to some extent. So, these new approaches can improve the performance of differentially private models.

\paragraph{Challenges:}
 The DP-based approaches are computationally costly. Additionally, as stated by the \cite{wu2022adaptive}, implementing the Selective Differential Privacy method requires knowledge of which items in the dataset contain private information. This requirement becomes prohibitively expensive, particularly for large-scale datasets. DP can provide privacy when there is a clear privacy border \cite{brown2020language} but can not provide privacy in some real-world scenarios like inference attack \cite{song2020information,zhou-etal-2022-textfusion}. Some studies suggest that DP can cause severe degradation of the model's performance \cite{jang2022knowledge}. 
 
\subsection{Private Representation Learning}
To avoid sharing sensitive information, instead of sharing plain text data directly, people can share representations\cite{lin2021fednlp}. However, these representations can still be transformed into the original text under a text reconstruction attack \cite{song2020information, pan2020privacy}. Private representation learning approaches can preserve inference privacy as they operate at the level of latent vector representations rather than modifying the texts themselves. These approaches can be applied to each word representation, making them indistinguishable or, at the targeted token, breaking the one-to-one relation between token representations and raw words and hiding private words.

The TextFusion approach by \citet{zhou-etal-2022-textfusion} does not change the basic architecture of the pre-trained language model but introduces a fusion predictor to determine which tokens should be fused. Then, the suitable token representations are fused in the privacy-preserving layer. This approach also tries to mislead the attacker by making the token representation more predictable to a different word with the closest Euclidean distance. The main goal of these fused representations is to make it challenging for privacy attackers to revert the token representations back to raw words, and the misleading training misleads the attacker for both fused and unfused representations.

Another framework, named TextObfuscator, proposed by \citet{zhou2023textobfuscator}, used the obfuscation technique to preserve privacy. The task is done in two steps; in the first step, each word is assigned to a corresponding prototype depending on semantic and task-related roles. In the second step for private representation training, the goal is to get a word representation and then make the representation close to its prototype by using the $\mathcal{L}_{\text{close}}$ in equation \ref{eq: Lclose}. Here, $p_{x_i}$ is the prototype of $x_i$ and the word representation is $H = \{h_i\}_{i=1}^{n}$.
\begin{align}
    \mathcal{L}_{\text{close}} = \frac{1}{2} \sum_{i=1}^{n} \lVert h_i - p_{x_i} \rVert_2^2
    \label{eq: Lclose}
\end{align}
Also, the distance between different prototypes is maintained to avoid collapse during training using prototype distance loss in equation \ref{eq: Laway}. Here, $n_p$ is the number of prototypes.
\begin{align}
    \mathcal{L}_{\text{away}} = \frac{2}{n_p(n_p - 1)} \sum_{i=1}^{n_p} \sum_{j=i+1}^{n_p} \lVert p_i - p_j \rVert_2^2
    \label{eq: Laway}
\end{align}

\paragraph{Benefits:}
According to some studies, DP-based approaches do not fully protect privacy under the text reconstruction attack during inference \cite{song2020information,zhou-etal-2022-textfusion}. Privatizing token representations during inference can overcome this problem. Also, these approaches can preserve privacy without substantially sacrificing performance.

\paragraph{Challenges:}
The TextFusion approach relies on getting the predictions for confident representations on the early layer for token classification, which will not be suitable for tasks requiring a large-scale fusion ratio. Also, for token classification, the fusion rate has a greater impact on the task performance.
The TextObfuscator requires more training steps compared with fine-tuning, resulting in increased computational cost. Also, the approach was designed for inference privacy and was not tested against other privacy attributes.

\subsection{Knowledge Unlearning approaches}
Machine unlearning is an approach to overcome data privacy issues in machine learning. It has been mostly used for preserving privacy image classification models. However, this unlearning approach has recently been adapted for forgetting targeted data in language models \cite{jang2022knowledge, yao2023large}. The unlearning task is more challenging for language models compared with classification tasks due to the larger output space($\sim$ a few image classes vs. a sequence of tokens that can each be classified into $V \in \mathbb{R} ^{\sim 50,000}$)

The knowledge unlearning approach for mitigating privacy by \citet{jang2022knowledge} proposes a method to unlearn a specific sequence of tokens for language models. The proposed approach negates the original training objective by training to maximize the negative log-likelihood loss of the token sequence. By going in the opposite direction of the traditional gradient descent, it reverts the effects learned from specific sequences of tokens. In the loss function described in Equation \ref{eq: unlearning loss}, $f_\theta$ denotes the model with parameters $\theta$, $x=\{x_1, x_2,\ldots,x_T\}$ is a sequence of tokens, and $p_\theta(x_t \mid x_{<t})$  represents the conditional probability of $x_t$ being the next token given the preceding tokens $x_{<t}$. The target is to maximize the loss $\mathcal{L}_{UL}$.
\begin{align}
    \mathcal{L}_{UL}(f_\theta, x) &= -\sum_{t=1}^{T} \log(p_\theta(x_t | x_{<t})) 
    \label{eq: unlearning loss}
\end{align}
Extraction Likelihood (EL) and Memorization Accuracy (MA) are used to quantify if a token sequence can be considered to be forgotten. As shown in equation \ref{eq: ELn}, given a token sequence $x=\{x_1,x_2,\ldots,x_T\}$ to a language model $f$ pretrained with parameters $\theta$, $EL_n$ is the total n-gram overlap of generated and target token sequences calculated by equation \ref{eq: Overlap}. Here, $ng(\cdot)$ is the n-grams of a given token sequence. If an n-gram $c$ is present in the n-grams of sequence $b$, then it is considered to overlap.
\begin{align}
    EL_n(x) = \frac{\sum_{t=1}^{T-n} OVERLAP_n(f_{\theta}(x_{<t}), x_{\geq t})}{T - n}
    \label{eq: ELn}
\end{align}
\begin{align}
    OVERLAP_n(a,b) = \frac{\sum_{c \in ng(a)} \mathds{1}\{c \in ng(b)\}}{|ng(a)|}
    \label{eq: Overlap}
\end{align}
Memorization Accuracy (MA) is calculated by equation \ref{eq: MA}, quantifying the memorization of a given token sequence by the language model. It is considered to be memorized if the token predicted by the LM at position $t$ matches the actual token $x_t$.
\begin{align} 
    MA(x) &= \frac{\sum_{t=1}^{T-1} \mathds{1}\{argmax(p_{\theta}(\cdot | x_{<t})) = x_t\}}{T - 1}
    \label{eq: MA}
\end{align}
A token sequence is considered forgotten if both $EL_n$ and $MA$ are lower than the average of token sequences from a validation corpus. In equations \ref{eq: Overlap} and \ref{eq: MA}, $\mathds{1}\{\cdot\}$ is the Indicator function i.e., $\mathds{1}\{True\} = 1$ and $\mathds{1}\{False\} = 0$.

Another gradient ascent loss-based unlearning approach has been proposed by \citet{yao2023large}. The main idea of this approach is that any task where the language model needs to forget the impact of certain training samples can be achieved by unlearning. To perform unlearning, it only requires the negative samples. Then, the gradient ascent loss is used to forget the negative samples.

\paragraph{Benefits:}
The unlearning approach is useful for making an LLM forget PII and copyright contents. It can preserve privacy under targeted extraction attacks. The approach is also cost-efficient as it doesn’t require re-training the whole language model \cite{jang2022knowledge}. It only updates the parameters for a few negative samples. Also, the knowledge unlearning approach causes minimal or negligible deterioration in the original LLM's performance. In some cases, it even results in notable enhancements in LLM performance. 

\paragraph{Challenges:}
A study by \citet{carlini2022privacy} suggests that machine unlearning can even degrade others’ privacy. According to \citet{carlini2022privacy}, if modifications occur to the underlying dataset, a data point that is presently safe from membership inference may later become vulnerable.

\subsection{Data preprocessing approaches}
The data preprocessing approach to preserving privacy in language models requires re-training an LLM with anonymised data. This section includes overviews of text anonymisation and deduplication, among the data preprocessing approaches.

According to \citet{lison-etal-2021-anonymisation}, the process of text anonymisation poses a significant challenge, even for human annotators, as it extends beyond simply identifying predetermined categories of entities. The anonymisation can be done via NLP approaches \cite{yermilov-etal-2023-privacy, papadopoulou-etal-2022-bootstrapping} or DP approaches \cite{yue-etal-2021-differential, chen-etal-2023-customized}. 
In NLP approaches, different techniques are employed to remove or mask privacy-sensitive information, such as de-identification and obfuscation. In de-identification, sensitive information is removed or masked by generic or anonymous identifiers \cite{yermilov-etal-2023-privacy, papadopoulou-etal-2022-bootstrapping}. Obfuscation is also done at the level of latent vector representations rather than modifying the text \cite{huang2020texthide, mosallanezhad2019deep}.
DP-based text anonymisation approaches use differential privacy principles during the anonymisation process. In SANTEXT, proposed by \citet{yue-etal-2021-differential}, the authors considered the entire document sensitive and sanitised it with a modified MLDP algorithm \cite{chatzikokolakis2013broadening}.

Some approaches involve deduplication of the training data to mitigate privacy risks in language models \cite{lee2021deduplicating, kandpal2022deduplicating}. Duplication is when a sequence of characters exactly matches with another sequence of characters. Language models have a tendency to regenerate duplicate sequences from the training data, and an adversary can use them to recover a training sequence. According to \citet{lee2021deduplicating}, the frequency of generating an N-length sequence by a model superlinearly increases with the increase of duplication of that sequence in the training data. 
During deduplication, two types of duplicate sequences can be considered: the exact substring duplication and the approximate or semantic duplication. But for the privacy-preserving task, the authors considered only the exact sequence as it matches the adversary's goal. When two examples, $x_i$ and $x_j$, share a sufficiently long substring, that substring is removed from one of them. 
This deduplication approach reduces the amount of training data generated by the models by reducing the regeneration of duplicate sequences.  

\paragraph{Benefits:}
The main advantages of data preprocessing approaches are that they can provide defence against extraction attacks to some extent and maintain overall model performance. Some approaches even improve performance, so these approaches are suitable for tasks where high performance is a critical factor. 
Also, the approach is less computationally demanding compared with DP-based approaches.

\paragraph{Challenges:}
Though privacy can be compromised by recovering approximate duplicates, deduplicating approximate duplicates is more challenging and future work can be done in this direction. Also, the preprocessing approach requires re-training the underlying language models each time it wants to stop a new sequence from regenerating, so it is not suitable for tasks where we need to update for only a few token sequences. According to recent studies \cite{brown2020language}, more than preprocessing approaches are required to remove privacy-sensitive data like bank passwords and medical records and provide weaker privacy protection against this type of information.  

\section{Key findings}
\label{sec: Key findings}
Our study on privacy-preserving methods for language modeling approaches has revealed several important insights. These are summarized below.
\begin{itemize} 
    \item Data preprocessing methods cannot fully provide privacy guarantees, which are insufficient for removing personally identifiable information like names, email addresses, and passwords.
    \item Data preprocessing and Knowledge Unlearning-based approaches do not harm the overall performances, but performance decreases for Differential Privacy-based approaches in some cases.
    \item Knowledge Unlearning does not require re-training the language model. It needs to perform parameter updates for a few tokens, which is faster than other approaches involving re-training.
     \item While other approaches do not fully protect privacy under the text reconstruction attack during inference, Private Representation Learning-based approaches help to preserve privacy in the inference phase.
     \item DP provides strong privacy, and recent models such as Adaptive DP and Selective DP improve the performance of differentially private models while maintaining privacy.
\end{itemize}

\section{Limitation and Future works}
\label{sec: Limitation and Future works}
In this section, we focus on some of the major limitations of the discussed approaches, as well as the limitations of this study. Based on these limitations, we also point out future possible research directions, e.g., expansion of methods to cover languages other than English, development of integrated privacy-preserving techniques, etc.

\paragraph{Language Coverage:}
Most current studies focus on the English language, but expanding research into privacy risks posed by language models working with other languages is important. Also, the recent progress in privacy-preserving language modelling should be reflected in language models operating in multilingual contexts. Future research should aim to bridge the gap by exploring privacy risks and adapting privacy-preserving methodologies to languages beyond English.

\paragraph{Methodological Limitations:}   
From the findings of this study, it becomes evident that no single method provides overall protection against diverse privacy risks. Data preprocessing-based approaches can not fail to remove personally identifiable information. Some studies \cite{song2020information,zhou-etal-2022-textfusion} suggested that the widely used DP-based approaches can not protect privacy at the inference phase. Also, the Knowledge Unlearning approach does not focus on inference attacks. Private Representation Learning approaches are focused on preserving privacy at the inference phase and can not give a guarantee about other privacy attributes. Some studies suggest combining multiple approaches to enhance privacy \cite{kandpal2022deduplicating}, but more work is needed to explore how multiple privacy methods can work together. Future research should also aim to develop integrated privacy-preserving strategies that address the limitations of existing methods, ensuring comprehensive protection against diverse privacy attacks.

\paragraph{Comparative Analysis and Unexplored Impacts of Knowledge Unlearning approach:}   
The existing Knowledge Unlearning studies lack a comprehensive comparison with more recent DP approaches, leaving an avenue for future research to explore and provide a better understanding of the effectiveness of these two privacy-preserving techniques.
Also, as discussed in section \ref{sec: Research Methods and Approaches}, Knowledge Unlearning can compromise other people's privacy. Further research is needed to investigate the impact of Knowledge Unlearning on other people's privacy.

\paragraph{Limitations of this work:} 
One limitation of this work is that it only focuses on four approaches: Private Representation Learning, Knowledge Unlearning, Data preprocessing, and Differential Privacy. Due to page limitations, we could not discuss some other available approaches, e.g., Federated Learning and Homomorphic Encryption. Also, this paper's findings are based on previous works, and this study does not include any case studies. These are left for future work.

\section{Conclusion}
\label{sec: Conclusion}
In recent years, LLMs are becoming an integral part of our lives. People are using LLMs with little or no knowledge of how many privacy risks these models pose. This study explains these privacy risks posed by the current LLMs and currently used approaches for mitigating these risks. This paper provides an overview of the approaches and their benefits and challenges. Privacy-preserving approaches in language modeling implement different strategies to mitigate privacy risks in language models. As described in this study, mitigating privacy risks is a difficult task, and various privacy attacks make the task more difficult. This study discusses four kinds of approaches: Private Representation Learning, Knowledge Unlearning, Data preprocessing, and Differential Privacy. All the approaches can preserve privacy against specific attacks, and none of them gives privacy against all types of attacks. The aim of this work was to understand the approaches and find out the gaps to help the community focus on the bigger unresolved questions. 

\printbibliography

\end{document}